\documentclass{bmvc2k}

\title{Searching for TrioNet: Combining Convolution with Local and Global Self-Attention}

\addauthor{Huaijin Pi}{hjpi@zju.edu.cn}{1}
\addauthor{Huiyu Wang}{hwang157@jhu.edu}{2}
\addauthor{Yingwei Li}{yingwei.li@jhu.edu}{2}
\addauthor{Zizhang Li}{	zzli@zju.edu.cn}{1}
\addauthor{Alan Yuille}{alan.l.yuille@gmail.com}{2}

\addinstitution{
 Zhejiang University\\
 Zhejiang, China
}
\addinstitution{
 Johns Hopkins University\\
 Baltimore, USA
}

\runninghead{H. Pi, H. Wang, Y. Li, Z. Li, A. Yuille}{TrioNet}

\def\ie{\emph{i.e}\bmvaOneDot}

\def\etal{\emph{et al}\bmvaOneDot}

\definecolor{vis_red}{rgb}{0.918,0.263,0.208}
\definecolor{vis_green}{rgb}{0.204,0.659,0.325}
\definecolor{vis_blue}{rgb}{0.259,0.522,0.957}
\definecolor{vis_yellow}{rgb}{0.984, 0.737, 0.016}

\makeatletter\renewcommand\paragraph{\@startsection{paragraph}{4}{\z@}
  {-.0em \@plus1ex \@minus.2ex}{-.5em}{\normalfont\normalsize\bfseries}}\makeatother
\newcommand{\figref}[1]{Fig.~\ref{#1}}
\newcommand{\tabref}[1]{Tab.~\ref{#1}}

\newcommand{\equref}[1]{Equ.~(\ref{#1})}

\usepackage{times}
\usepackage{epsfig}
\usepackage{graphicx}
\usepackage{amsmath}
\usepackage{amssymb}

\usepackage{dsfont}
\usepackage{bm}  
\usepackage{booktabs}
\usepackage{xspace}  %
\usepackage{multirow}
\usepackage{colortbl}
\usepackage{enumitem}
\usepackage{placeins}
\usepackage{footnote}
\usepackage{multicol}
\usepackage{graphbox}
\usepackage{float}
\usepackage{placeins}
\usepackage{mathrsfs} 

\newfloat{figtab}{htb}{fgtb}
\makeatletter
  \newcommand\figcaption{\def\@captype{figure}\caption}
  \newcommand\tabcaption{\def\@captype{table}\caption}
\makeatother

\usepackage{algorithmic}
\usepackage{algorithm}
\definecolor{commentcolor}{RGB}{110,154,155}   
\newcommand{\PyComment}[1]{\ttfamily\textcolor{commentcolor}{\# #1}}  
\newcommand{\PyCode}[1]{\ttfamily\textcolor{black}{#1}} 

\usepackage{hyperref}

\begin{document}
\maketitle
\vspace{-3ex}
\begin{abstract}
Recently, self-attention operators have shown superior performance as a stand-alone building block for vision models. However, existing self-attention models are often hand-designed, modified from CNNs, and obtained by stacking one operator only. A wider range of architecture space which combines different self-attention operators and convolution is rarely explored. In this paper, we explore this novel architecture space with weight-sharing Neural Architecture Search (NAS) algorithms. The result architecture is named TrioNet for combining convolution, local self-attention, and global~(axial) self-attention operators. In order to effectively search in this huge architecture space, we propose Hierarchical Sampling for better training of the supernet. In addition, we propose a novel weight-sharing strategy, Multi-head Sharing, specifically for multi-head self-attention operators. Our searched TrioNet that combines self-attention and convolution outperforms all stand-alone models with fewer FLOPs on ImageNet classification where self-attention performs better than convolution. Furthermore, on various small datasets, we observe inferior performance for self-attention models, but our TrioNet is still able to match the best operator, convolution in this case. Our code is available at \href{https://github.com/phj128/TrioNet}{https://github.com/phj128/TrioNet}.
\end{abstract}

\vspace{-4ex}
\section{Introduction}
\vspace{-1.5ex}
\label{sec:intro}

Convolution is one of the most commonly used operators for computer vision applications \cite{he2016deep, ren2015faster, he2017mask}. However, recent studies \cite{hu2019local, parmar2019stand, Zhao2020ExploringSF, wang2020axial, dosovitskiy2021an} suggest that the convolution operator is unnecessary for computer vision tasks, and that the self-attention module \cite{vaswani2017attention} might be a better alternative. For example, Axial-DeepLab \cite{wang2020axial} built stand-alone self-attention models for image classification and segmentation. Although these attention-based models show promising results to take the place of convolution-based models, these attention-based models are all human-designed \cite{parmar2019stand, wang2020axial}, making it challenging to achieve optimal results on new dataset, tasks, or computation budgets.

\begin{figure}[t]
\centering
\includegraphics[width=0.8\textwidth]{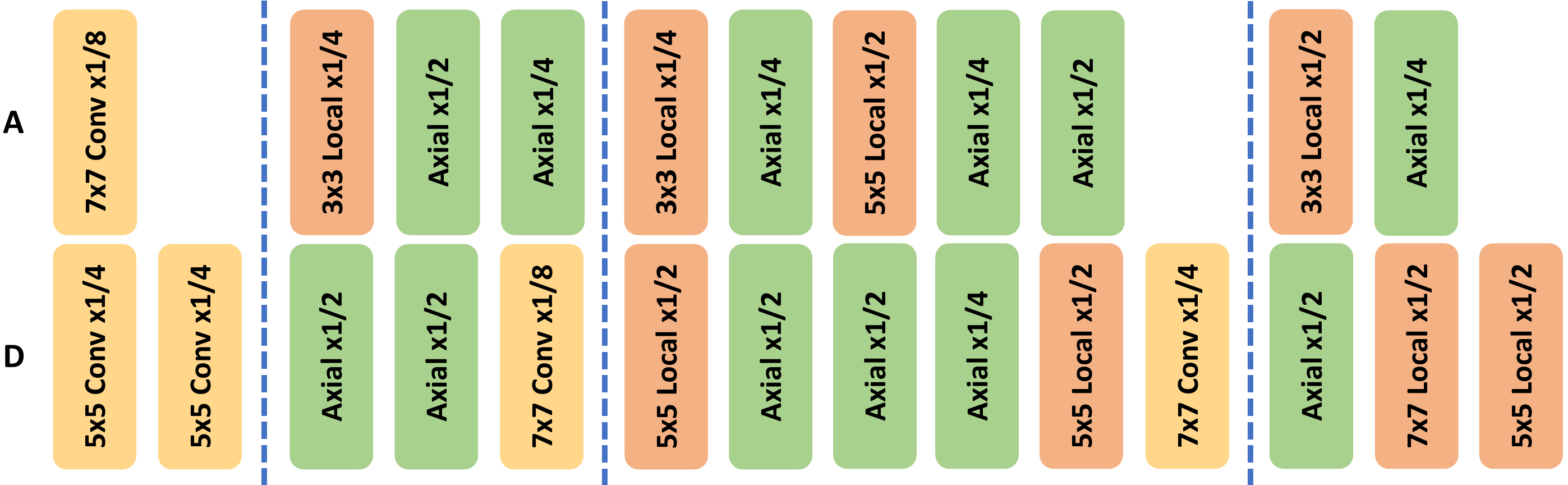}
\vspace{-2ex}
\caption{Searched TrioNet A and D. Conv, Local and Axial denotes convolution, local-attention and axial-attention. $\times$1/2, $\times$1/4, $\times$1/8 are expansion rates. TrioNet consistently uses convolutions at low level and self-attention at high level.
}
\vspace{-3.5ex}
\label{fig:searchedmodel}
\end{figure}

Given a target dataset, task, and computation budget, Neural Architecture Search (NAS) \cite{2017rlnas, Zoph2018LearningTA, Real2019RegularizedEF, liu2018darts, liu2018hierarchical, Pham2018EfficientNA, wu2019fbnet, efficientnet} is an efficient technology to automatically find desirable architectures with marginal human labor. Due to its effectiveness, previous works \cite{2017rlnas, liu2018darts, Pham2018EfficientNA, Cai2020Once-for-All:, Yu2020BigNASSU} have successfully applied NAS to different computer vision tasks, including object detection \cite{Wang2020NASFCOSFN, Ghiasi2019NASFPNLS}, semantic segmentation \cite{Liu2019AutoDeepLabHN}, medical image analysis \cite{Zhu2019VNASNA}, and video understanding \cite{Wang2020PVNASPN}. However, most of the searched models for these computer vision tasks are built upon convolutional neural network (CNN) backbones \cite{howard2017mobilenets, sandler2018mobilenetv2, howard2019searching}. 

Previous convolution-based NAS methods usually consider attention modules as plugins \cite{Li2020NeuralAS}. During the search procedure, the search algorithm needs to decide whether an attention module should be appended after each convolution layer. This strategy results in the searched network architecture majorly consisting of convolution operators, causing a failure to discover the potentially stronger attention-based models.

Therefore, in this paper, we study how to search for combined vision models which could be fully convolutional or fully self-attentional~\cite{parmar2019stand, wang2020axial}. Specifically, we search for TrioNets where all three operators (local-attention~\cite{parmar2019stand}, axial-attention~\cite{wang2020axial}, and convolution) are considered equally important and compete with each other. Therefore, self-attention is no longer an extra plugin but a primary operator in our search space. Seemingly including the self-attention module into the search space could achieve our goal, we observe it is difficult to apply the commonly used weight-sharing NAS methods~\cite{cai2018proxylessnas,Cai2020Once-for-All:,Yu2020BigNASSU} due to two issues.

One issue is that self-attention modules and convolution blocks make our search space much more complicated and imbalanced than convolution-only spaces~\cite{cai2018proxylessnas, Cai2020Once-for-All:, Yu2020BigNASSU}. For instance, convolution usually contains kernel size and width to search while self-attention have more options like query, key and value channels, spatial extent and multi-head numbers. The self-attention operators correspond to much larger search spaces than convolution, with the same network depth. This imbalance of self-attention and convolution's search space makes the training of supernets intractable. Therefore, we propose Hierarchical Sampling, which samples the operator first uniformly before sampling other architecture options. This sampling rule suits our setting better because it ensures an equal chance for each operator to be trained in the supernet, alleviating the bias of search space size.

The other issue is that the multi-head design of attention operators poses a new challenge for weight-sharing \cite{Cai2020Once-for-All:, Yu2020BigNASSU} NAS algorithms. Current weight-sharing strategy \cite{Stamoulis2019SinglePathND, Cai2020Once-for-All:, Yu2020BigNASSU} always shares the first few channels of a full weight matrix to construct the weight for small models. However, in self-attention modules, the channels are split into multi-head groups to capture different dependencies \cite{vaswani2017attention}. The current weight-sharing strategy ignores the multi-head structure in the weights and allocates the same channel to different heads depending on the sampled multi-head groups and channels, forcing the same channel to capture different types of dependencies at the same time. We hypothesize and verify by experiments that such weight-sharing is harmful to the training of supernets. Instead, we share our model weights only if they belong to the same head in multi-head self-attention. This dedicated strategy is named Multi-Head Sharing strategy.



We evaluate our TrioNet on ImageNet \cite{krizhevsky2012imagenet} dataset and various small datasets \cite{stanfordcars, fgvc, cub, caltech101, 102flowers}. The TrioNet architectures found on ImageNet are shown in \figref{fig:searchedmodel}. We observe that TrioNets outperform stand-alone convolution~\cite{he2016deep}, local-attention~\cite{parmar2019stand} and axial-attention \cite{wang2020axial} models with fewer FLOPs on ImageNet where self-attention performs better than convolution. On small datasets where self-attention models perform inferior to convolution, our TrioNet is still able to matche the best operator with fewer FLOPs on average.

To summarize, our contributions are four-fold:
(1) We regard self-attention and convolution as equally important basic operators and propose a new search space that contains both stand-alone self-attention models and convolution models.
(2) In order to train a supernet on our highly imbalanced search space, we adopt Hierarchical Sampling rules to balance the training of convolution and self-attention operators.
(3) A Multi-Head Sharing strategy is specifically designed for sharing weights in multi-head self-attention operators.
(4) Our searched TrioNet reduces computation costs and improves results on ImageNet classification, compared with hand-designed stand-alone networks. The same phenomenon is observed when TrioNets are searched on small datasets as well.

\vspace{-4ex}
\section{Related work}
\vspace{-1.5ex}

\paragraph{Self-attention.}
Self-attention was firstly proposed for NLP tasks \cite{bahdanau2014neural, wu2016google, vaswani2017attention}. 
People then successfully apply self-attention module to many computer vision tasks \cite{wang2018non,bello2019attention,chen20182,huang2019ccnet, zhu2019asymmetric, Cao2019GCNetNN}.
More recently, it has been shown that self-attention can be used to replace convolution in vision models. Hu \etal~\cite{hu2019local} and Local self-attention~\cite{parmar2019stand} build the whole model with self-attention restricted to a local patch. SAN \cite{Zhao2020ExploringSF} explores a boarder self-attention formulation. Axial-DeepLab \cite{wang2020axial} extends local self-attention to global self-attention with axial self-attention blocks. Later, ViT \cite{dosovitskiy2021an} approaches image classification with the original NLP transformer architecture. Variants \cite{Wu2021CvTIC, Han2021TransformerIT, Liu2021SwinTH, Yuan2021TokenstoTokenVT} of ViT propose a few hand-designed ways of applying local constraints to the global self-attention in ViT. In this paper, we automatically search for an architecture in a combined space that includes convolution, local self-attention~\cite{parmar2019stand}, and axial global self-attention~\cite{wang2020axial}.

\paragraph{Neural Architecture Search.} Neural Architecture Search (NAS) was proposed to automate architecture design \cite{2017rlnas}. Early NAS methods usually sample many architectures and train them from scratch for picking up a good architecture, leading to large computational overhead \cite{2017rlnas, Zoph2018LearningTA, Real2019RegularizedEF, liu2018hierarchical, efficientnet, Cai2020Once-for-All:, Yu2020BigNASSU, sahni2021compofa}. More recently, people develop an one-shot pipeline by training a single over-parameterized network and sampling architectures within it to avoid the expensive train-from-scratch procedure \cite{2018understand, brock2018smash, liu2018darts, Pham2018EfficientNA, cai2018proxylessnas, Yu2019NetworkSB, Hu2020DSNASDN, Guo2020SinglePO, chu2021fairnas}. 
However, the search space of these methods is restricted in the MobileNet-like space \cite{howard2017mobilenets, sandler2018mobilenetv2, howard2019searching}. There are also some attempts to search for self-attention vision models \cite{Li2020NeuralAS, Wang2020AttentionNASSA, Li2021BossNASEH}. 
These works mainly focus on the single block design \cite{Wang2020AttentionNASSA} and positions \cite{Li2020NeuralAS}, or searches in a small space \cite{Li2021BossNASEH}. In this paper, self-attention and convolution are considered equally in our search space. 

\vspace{-4ex}
\section{Method}
\vspace{-1.5ex}

In this section, we first define our operator-level search space that contains both convolution and self-attention. Next, we discuss our one-shot architecture-level search algorithm that trains a supernet. Finally, we present our proposed Hierarchical Sampling (HS) and Multi-Head Sharing (MHS) that helps training the supernet.

\vspace{-3ex}
\subsection{Operator-Level Search Space} \label{operator}
\vspace{-1ex}
Convolution is usually the default and the only operator for a NAS space \cite{2017rlnas, efficientnet, cai2018proxylessnas, Cai2020Once-for-All:, Yu2020BigNASSU}. In this paper, however, we introduce self-attention operators into our operator-level space and search for the optimal combination of convolution and self-attention operators \cite{vaswani2017attention}. Specifically, we include efficient self-attention operators that can be used as a stand-alone operator for a network. 
We use axial-attention \cite{wang2020axial} instead of fully connected 2D self-attention \cite{dosovitskiy2021an} as an instantiation of global self-attention for computational efficiency.

\paragraph{Local Self-Attention.} Local self-attention~\cite{parmar2019stand} limits its receptive field to a local window. Given an input feature map $x \in \mathbb{R}^{h \times w \times d_{in}}$ with height $h$, width $w$, and channels $d_{in}$, the output at position $o=(i,j)$, $y_{o} \in \mathbb{R}^{d_{out}}$, is computed by pooling over the projected input as:
\vspace{-1ex}
\begin{equation}
\vspace{-1.5ex}
\label{eqn:standalone}
    y_{o} = \sum_{p \in \mathcal{N}_{m \times m}(o)} \text{softmax}_{p}(q_{o}^T k_{p} + q_{o}^T r_{p-o}) v_{p}
\end{equation}
where $\mathcal{N}_{m \times m}(o)$ is the local $m \times m$ square region centered around location $o=(i,j)$. Queries $q_{o}=W_Q x_{o}$, keys $k_{o}=W_K x_{o}$, values $v_{o}=W_V x_{o}$ are all linear projections of the input $x_{o}~\forall o \in \mathcal{N}$. $W_Q, W_K \in \mathbb{R}^{d_{q} \times d_{in}}$. $W_V \in \mathbb{R}^{d_{out} \times d_{in}}$ are all learnable matrices. The relative positional encodings $r_{p-o} \in \mathbb{R}^{d_{q}}$ are also learnable vectors and the inner product $q_{o}^T r_{p-o}$ measures the compatibility from location $p=(a,b)$ to location $o=(i,j)$. In practice, this single-head attention in \equref{eqn:standalone} is extended to multi-head attention to capture a mixture of affinities~\cite{vaswani2017attention}. 
In particular, multi-head attention is computed by applying $N$ single-head attentions in parallel on $x_{o}$ (with different learnable matrices $W_Q^n, W_K^n, W_V^n, \forall n \in \{1, 2, \dots, N\}$ for the $n$-th head), and then obtaining the final output $z_{o}$ by concatenating the results from each head, \ie, $z_{o}=\text{concat}_n(y^n_{o})$.

The choice of local window size $m$ significantly affects model performance and computational cost. Besides, it is not clear how many local self-attention layers should be used for each stage or how to select a window size for each individual layer. For these reasons, we include local self-attention in our search space to find a good configuration.

\paragraph{Axial Self-Attention.} Axial-attention~\cite{wang2020axial,huang2019ccnet,ho2019axial} captures global relations by factorizing a 2D self-attention into two consecutive 1D self-attention operators, one on the height-axis, followed by one on the width axis 
Both of the axial-attention layers adopt the multi-head attention mechanism, as described above. Note that we do not use the PS-attention~\cite{wang2020axial} or BN~\cite{ioffe2015batch} layers for fair comparison with local-attention~\cite{parmar2019stand} and faster implementation.

Despite capturing global contexts, axial-attention is less effective to model local relations if two local pixels do not belong to the same axis. In this case, combining axial-attention with convolution or local-attention is plausible. So we study the combination in this paper by including axial-attention into our operator-level search space.

\vspace{-3ex}
\subsection{Architecture-Level Search Space}
\vspace{-1ex}

\begin{table*}[bt]
\footnotesize
    \centering
    \begin{tabular}{c c c c}
        \hline
        Operator & Convolution & Local Attention~\cite{parmar2019stand} & Axial Attention~\cite{wang2020axial}\\
        \hline
        Expansion rates & 1/8, 1/4 & 1/4, 1/2 & 1/4, 1/2 \\
        Kernel size & 3, 5, 7 & 3, 5, 7 & - \\
        Query (key) channel rates & - & 1/2, 1 & 1/2, 1 \\
        Value channels rates & - & 1/2, 1 & 1/2, 1\\
        Number of heads & - & 4, 8 & 4, 8 \\
        \hline
        Total choices (cardinality) & 6 & 48 & 16 \\
        \hline
    \end{tabular}
    \caption{The imbalanced search space for each operator.}
    \label{tab:space}
    \vspace{-4ex}
\end{table*}

Similar to hand-designed self-attention models, Local-attention~\cite{parmar2019stand} and Axial-DeepLab~\cite{wang2020axial}, we employ a ResNet-like \cite{he2016deep} model family to construct our architecture-level search space. Specifically, we replace all 3$\times$3 convolutions by our operator-level search space that contains convolution, local-attention, and axial-attention.

\tabref{tab:space} summarizes our search space for each block. Different from a cell-level search space in the NAS literature~\cite{Cai2020Once-for-All:, Yu2020BigNASSU}, our spatial operators are not shared in each residual block. As a result, our search space allows a flexible combination of different operators at each layer.
This space also includes pure convolutional \cite{he2016deep}, pure local-attention \cite{parmar2019stand}, and pure axial-attention\cite{wang2020axial} ResNets. Our search space contains roughly $7.2 \times 10^{25}$ models in total, 3.6 M times larger than that of OFA \cite{Cai2020Once-for-All:} ($2 \times 10^{19}$).

\vspace{-3ex}
\subsection{Searching Pipeline}
\vspace{-1ex}

\begin{figure}[t]
\centering
\includegraphics[width=0.7\textwidth]{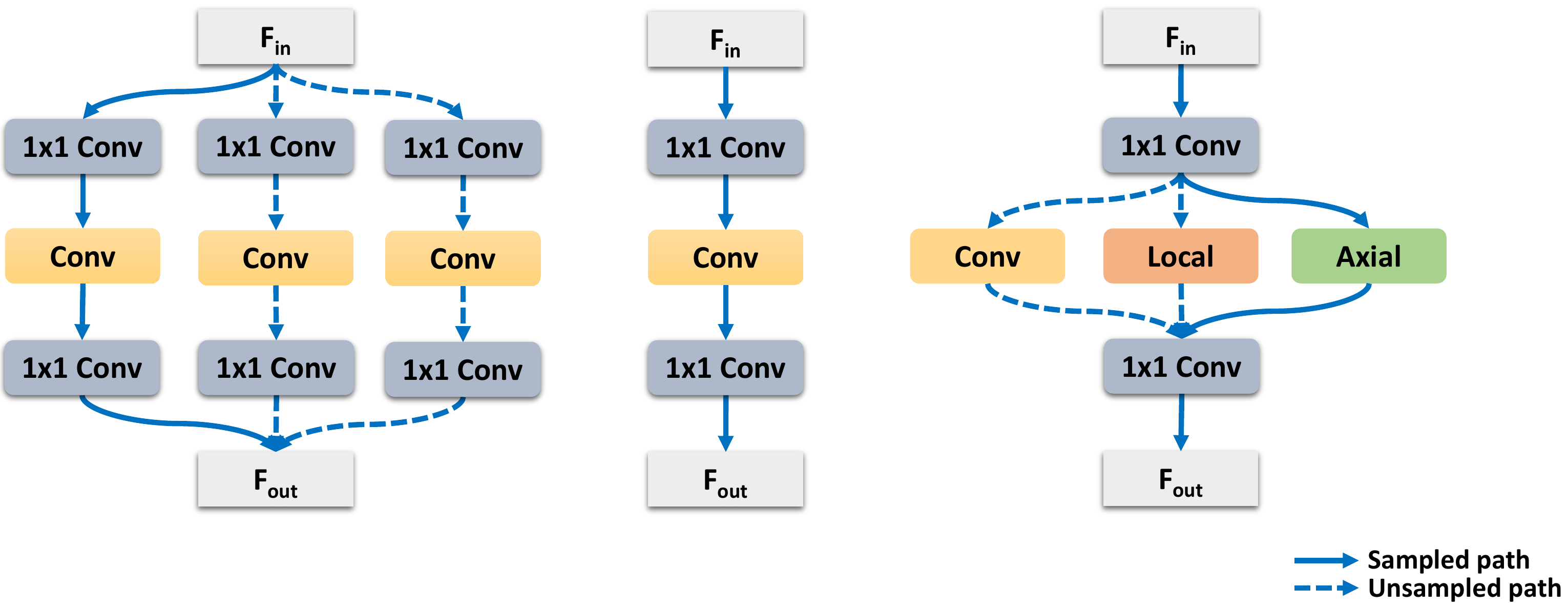}
\vspace{-2ex}
\caption{{\bf Supernet candidate sampling.} Left: ProxylessNAS-style \cite{cai2018proxylessnas}. Middle: OFA-style \cite{Cai2020Once-for-All:}. Right: our TrioNet with shared projection layers and three separate parallel operators.}
\vspace{-4ex}
\label{fig:blockmodel}
\end{figure}

Given this search space, we employ the one-shot NAS pipeline \cite{Cai2020Once-for-All:, Yu2020BigNASSU}, where the entire search space is built as a weight-sharing supernet. The typical strategy~\cite{cai2018proxylessnas, Stamoulis2019SinglePathND} of most one-shot NAS works is shown in \figref{fig:blockmodel} a). A MobileNet-like~\cite{howard2017mobilenets, sandler2018mobilenetv2, howard2019searching} supernet is built with blocks that contain several parallel candidates with different channels and receptive fields. Consequently, the supernet needs a lot of parameters to hold the large search space, making it inapplicable to our huge search space (\tabref{tab:space}). OFA and BigNAS \cite{Cai2020Once-for-All:, Yu2020BigNASSU} propose to share candidate weights in a single block, as shown in \figref{fig:blockmodel} b). This allows a much larger search space with manageable parameters but it still limited one operator only. In our case, different spatial operators can not share all the parameters.

Based on these works \cite{cai2018proxylessnas, Cai2020Once-for-All:, Yu2020BigNASSU}, we make a step forward. We insert local-~\cite{parmar2019stand} and axial-~\cite{wang2020axial} attention into the single block parallel with the spatial convolution as the primary spatial operator. These parallel spatial operators share the same projection layers and increase the flexibility of the supernet without introducing many parameters. 
Like \cite{Cai2020Once-for-All:, Yu2020BigNASSU}, we can formulate the problem as 
\vspace{-1.5ex}
\begin{equation}
\vspace{-2ex}
\label{eqn:training}
    \min _{W_{o}} \mathbb{E}_{\alpha \sim \Gamma(\mathcal{A})}[\mathcal{L}_{\text {train}}\left(C\left(W_{o}, \operatorname{\alpha}\right)\right)]
\end{equation}
where $\left(C\left(W_{o}, \operatorname{\alpha}\right)\right)$ denotes that the selection of the parameters from the weights of the supernet $W_{o}$ with architecture configuration $\alpha$, $\Gamma(\mathcal{A})$ is the sampling distribution of $\alpha \in \mathcal{A}$ and $\mathcal{L}_{\text {train}}$ denotes the loss on the training dataset. After training the supernet, an evolutionary algorithm is used to obtain the optimal model ${\alpha}_{o}$ on the evolutionary search validation dataset with the goal as 
\vspace{-1ex}
\begin{equation}
\vspace{-1ex}
\label{eqn:search}
    {\alpha}_{o} = \underset{\alpha}{\operatorname{argmin}}\mathcal{L}_{\text {val }}\left(C\left(W_{o}, \operatorname{\alpha}\right)\right)
\end{equation}
Finally, we retrain the searched model from scratch on the whole training set and validate it on the validation set. 

Next, we discuss two issues of the default supernet training pipeline and our solutions.

\vspace{-3ex}
\subsection{Hierarchical Sampling} 
\vspace{-1ex}
\label{samplingstrategy}

Previous work~\cite{wang-etal-2020-hat} trains supernets by sampling candidates uniformly. However, as shown in \tabref{tab:space}, our search space candidates are highly biased towards the local-attention operator, leading to an imbalanced training of the operators and worse performance of the searched model. Therefore, we propose Hierarchical Sampling (HS). For each block, we first sample the spatial operator uniformly. Given the operator, we then randomly sample a candidate from the operator space.
We can formulate this as 
\vspace{-1.5ex}
\begin{equation}
\vspace{-1.5ex}
\label{eqn:hs}
    \alpha = (\beta, \theta), \beta \sim \mathcal{U}(\mathcal{B}), \theta \sim \mathcal{U}(\Theta)
\end{equation}
where the architecture $\alpha$ can be expressed as operators configuration $\beta \in \mathcal{B}$ sampled from \tabref{tab:space} (the first row) and weights configuration $\theta \in \Theta$ sampled from \tabref{tab:space} (from the second to the sixth row) with uniform distribution $\mathcal{U}$. In this way, we ensure an equal sampling chance for all operators. 

In addition, we notice that the training is biased towards middle-sized models. We attributes this to the fact that the search space is full of middle-sized models and thus random sampling trains mostly middle-sized models. To address this issue, we use Sandwich rule~\cite{Yu2019UniversallySN, Yu2020BigNASSU} that samples the smallest candidate, the biggest candidate and 2 random candidates. Sandwich rule is adopted in our Hierarchical Sampling after the operator for each block is selected.

\vspace{-3ex}
\subsection{Multi-Head Sharing}
\vspace{-1ex}


\begin{figure}[tb]
  \hspace{2ex}
  \begin{minipage}[c]{0.5\textwidth}
    \includegraphics[width=0.95\linewidth]{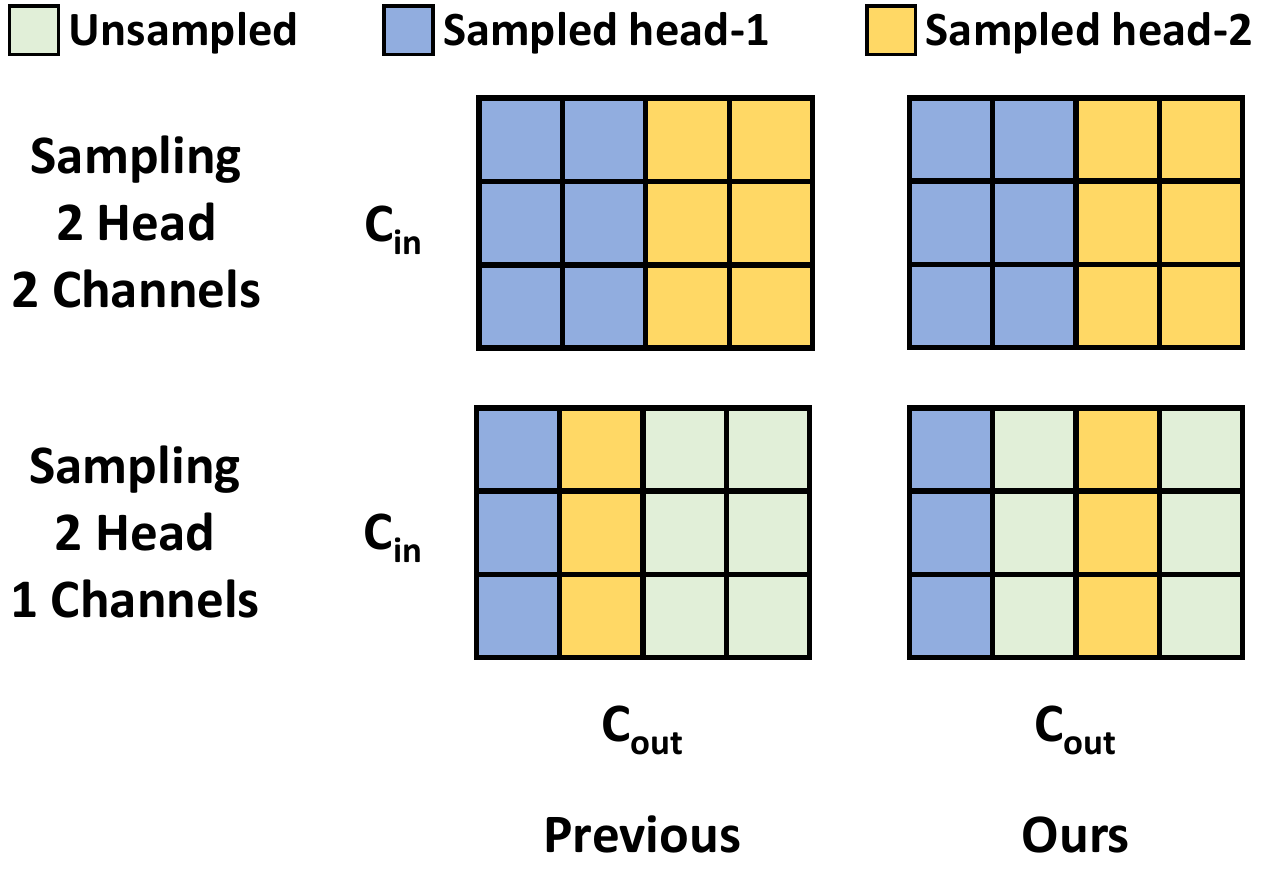}
  \end{minipage}  ~~
  \begin{minipage}[c]{0.37\textwidth}
    \caption{{\bf Multi-Head Sharing.} Previous: The default weight-sharing samples the second output channel sometimes for head 1 (top) and sometimes for head 2 (bottom). Ours: Our proposed Multi-Head Sharing samples all channels consistently according to the multi-head structure.}  \label{fig:weightsharing}
  \end{minipage}
  \vspace{-1em}
\end{figure}

The weight-sharing strategy for convolution has been well studied. Previous works \cite{Stamoulis2019SinglePathND, Yu2020BigNASSU, Cai2020Once-for-All:, Bender2020CanWS} share common parts of a convolution weight for different kernel sizes and channels. However, this strategy cannot be simply applied to our search space with multi-head self-attention operators. The varying number of heads makes sharing channels more complicated than convolution. In multi-head attention, the channels are split into multi-head groups to capture different dependencies~\cite{vaswani2017attention}. Thus, as shown in \figref{fig:weightsharing}, the default weight-sharing strategy allocates the same channels for different multi-head groups, which is harmful to the supernet training. To deal with this issue, we propose Multi-Head Sharing (MHS). As shown in \figref{fig:weightsharing}, we take into account the multi-head structure and first split all output channels into number-of-head groups. Then, we share channel weights only if they belong to the same head in multi-head self-attention. We show the selection procedure with PyTorch-like \cite{paszke2019pytorch} pseudo code in Alg. \ref{algo:mhs}. In this way, we ensure that different channel groups do not interfere with each other.

\vspace{-1em}
\begin{algorithm}[H]
\caption{Pseudo code of Multi-Head Sharing in a PyTorch-like style}
\algsetup{linenosize=\tiny}
\scriptsize
    \PyComment{w: weight matrix, c\_in: in channels}\\
    \PyComment{c\_out: max out channels, s\_c\_out: sampled out channels, n: multi-head number} \\
    \PyCode{h\_c = s\_c\_out / n}\\
    \PyCode{w = w.reshape(n, -1, c\_in)} \\
    \PyCode{w = w[:, :h\_c, :]} \PyComment{h\_c, n, c\_in}\\
    \PyCode{w = w.reshape(s\_c\_out, c\_in)} \\
    \PyCode{return w} \\
    \label{algo:mhs}
    \vspace{-1em}
\end{algorithm}
\vspace{-2em}

\vspace{-2ex}
\section{Experiments}
\vspace{-0.5ex}

Our main experiments are conducted on ImageNet \cite{krizhevsky2012imagenet} dataset. We also provide detailed ablation studies on the proposed Hierarchical Sampling and Multi-Head Sharing. Finally, we evaluate the adaptation of our searching algorithm on various small classification datasets.

\vspace{-3ex}
\subsection{ImageNet Classification}
\vspace{-1ex}

\tabref{tab:cls} shows the main results of our searched models with different FLOPs constraints. The searched models are compared with stand-alone convolution \cite{he2016deep},  stand-alone local self-attention~\cite{parmar2019stand} and stand-alone axial-attention~\cite{wang2020axial} models. As shown in \figref{fig:acc}, with 2B FLOPs budget, our TrioNet-A achieves 74.8\% accuracy and outperforms ResNet with 57.4\% less computation. With 3B FLOPs budget, our TrioNet-B achieves 75.9\% accuracy, which outperforms stand-alone local self-attention and axial-attention respectively with 33.3\% and 14.3\% fewer FLOPs. These results show that our searched TrioNet is able to outperform all hand-designed single-operator networks in terms of computation efficiency. In addition to our main focus in the low computation regime, we also evaluate models in the high computation regime. With 4B and 4.7B FLOPs budgets, our TrioNet-C and TrioNet-D achieve comparable performance-FLOPs trade-offs with stand-alone axial-attention and still outperforms fully convolution \cite{he2016deep} or local self-attention~\cite{parmar2019stand} methods with fewer FLOPs. Compared with OFA \cite{Cai2020Once-for-All:} under the same training settings as ours, TrioNet outperforms OFA with 1.1\%, 0.2\%, 0.2\% and 0.3\% accuracy with 16.7\%, 11.8\%, 13.0\% and 2.1\% less computation. Note that the larger models (TrioNet-C and TrioNet-D) are already close to the limit of our architecture space, which might lead to performance degrade. If large models instead of lightweight models are desired, our TrioNet searching pipeline can be directly extended to a high computation search space as well.

\vspace{-3ex}
\subsection{Ablation Studies}
\vspace{-1ex}
\begin{figtab}[t]
\begin{minipage}[c]{.49\linewidth}
\includegraphics[width=0.9\textwidth]{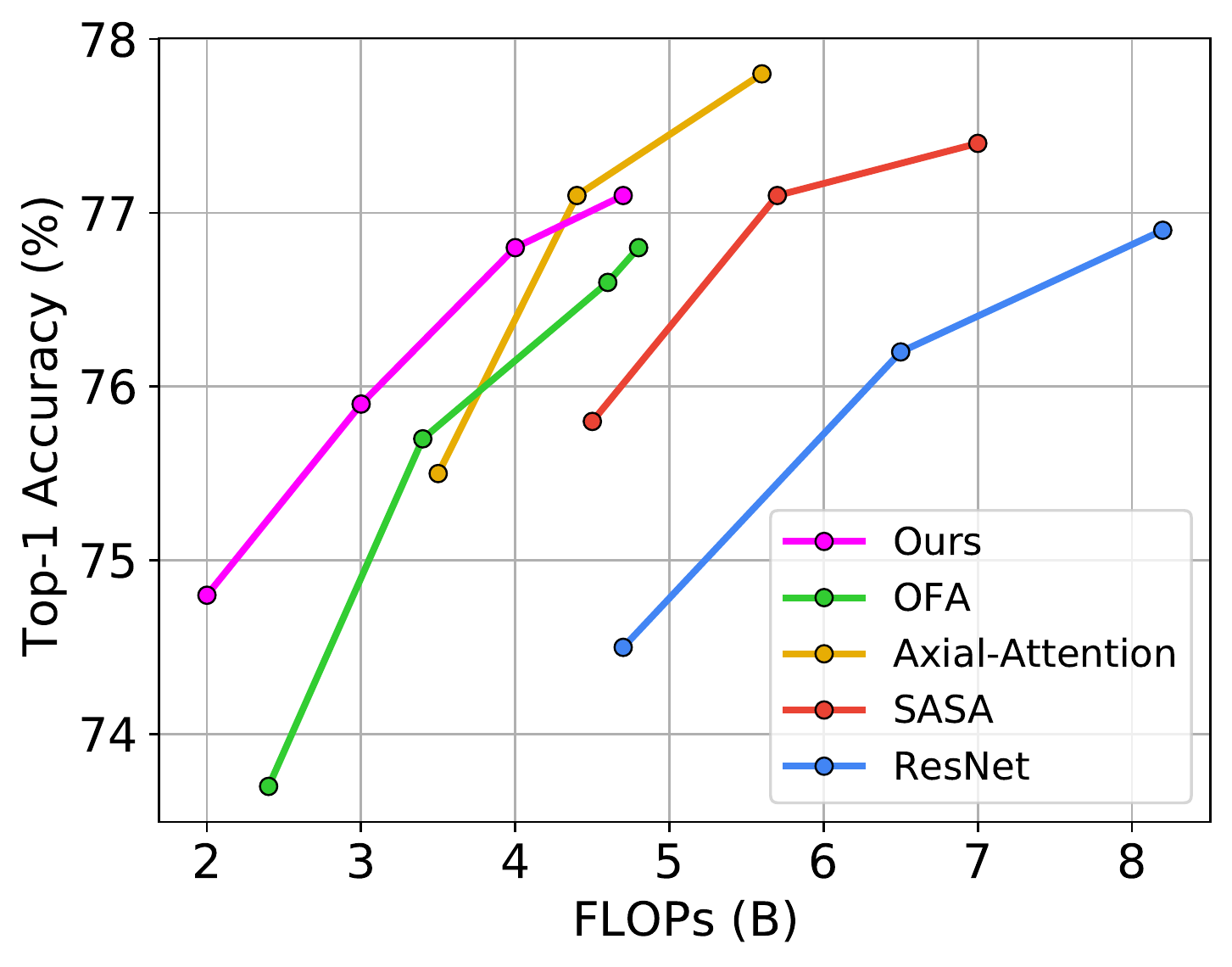}
\figcaption{Comparing Top-1 accuracy vs. FLOPs on ImageNet classification.}
\label{fig:acc}
\end{minipage}
\begin{minipage}[c]{.49\linewidth}
\footnotesize
    \centering
    \begin{tabular}{c cc c}
        \hline
        \textbf{Model} & \textbf{Params}  & \textbf{FLOPs}  & \textbf{Top-1} \\
        \hline
        ResNet-26 \cite{he2016deep, parmar2019stand} & 13.7M & 4.7B & 74.5 \\
        ResNet-38 \cite{he2016deep, parmar2019stand} & 19.6M & 6.5B & 76.2 \\
        ResNet-50 \cite{he2016deep, parmar2019stand} & 25.6M & 8.2B & 76.9 \\
        OFA-A \cite{Cai2020Once-for-All:} \footnotemark[1]\global\let\saved  & 9.6M & 2.4B & 73.7 \\
        OFA-B \cite{Cai2020Once-for-All:} \footnotemark[1]\global\let\saved & 14.4M & 3.4B & 75.7 \\
        OFA-C \cite{Cai2020Once-for-All:} \footnotemark[1]\global\let\saved & 19.2M & 4.6B & 76.6 \\
        OFA-D \cite{Cai2020Once-for-All:} \footnotemark[1]\global\let\saved & 20.7M & 4.8B & 76.8 \\
        \hline
        Local-26 \cite{parmar2019stand} & 10.3M & 4.5B & 75.8 \\
        Local-38 \cite{parmar2019stand} & 14.1M & 5.7B & 77.1 \\
        Local-50 \cite{parmar2019stand} & 18.0M & 7.0B & 77.4 \\
        Axial-26 \cite{wang2020axial} \footnotemark[2]\global\let\saved & 5.9M & 3.5B & 75.5 \\
        Axial-38 \cite{wang2020axial} \footnotemark[2]\global\let\saved & 8.7M & 4.4B & 77.1 \\
        Axial-50 \cite{wang2020axial} \footnotemark[2]\global\let\saved & 12.4M & 5.6B & 77.8 \\
        \hline
        TrioNet-A & 5.1M & 2.0B & 74.8 \\
        TrioNet-B & 9.4M & 3.0B & 75.9 \\
        TrioNet-C & 10.6M & 4.0B & 76.8 \\
        TrioNet-D & 10.9M & 4.7B & 77.1 \\
        \hline
    \end{tabular}
    \tabcaption{Results on ImageNet classification.}
    \label{tab:cls}
\end{minipage}
\vspace{-3ex}
\end{figtab}

\footnotetext[1]{Our retrained results.}
\footnotetext[2]{Our retrained models removing some BN and positional encoding.}

In this subsection, we provide more insights by ablating each of our proposed components separately. The experiments are also performed on ImageNet~\cite{krizhevsky2012imagenet}. We monitor the training process of the supernets by directly evaluating the largest possible model for each operator.

\begin{figure}[t]
\begin{tabular}{cc}
\centering
\quad \; \includegraphics[width=0.35\textwidth]{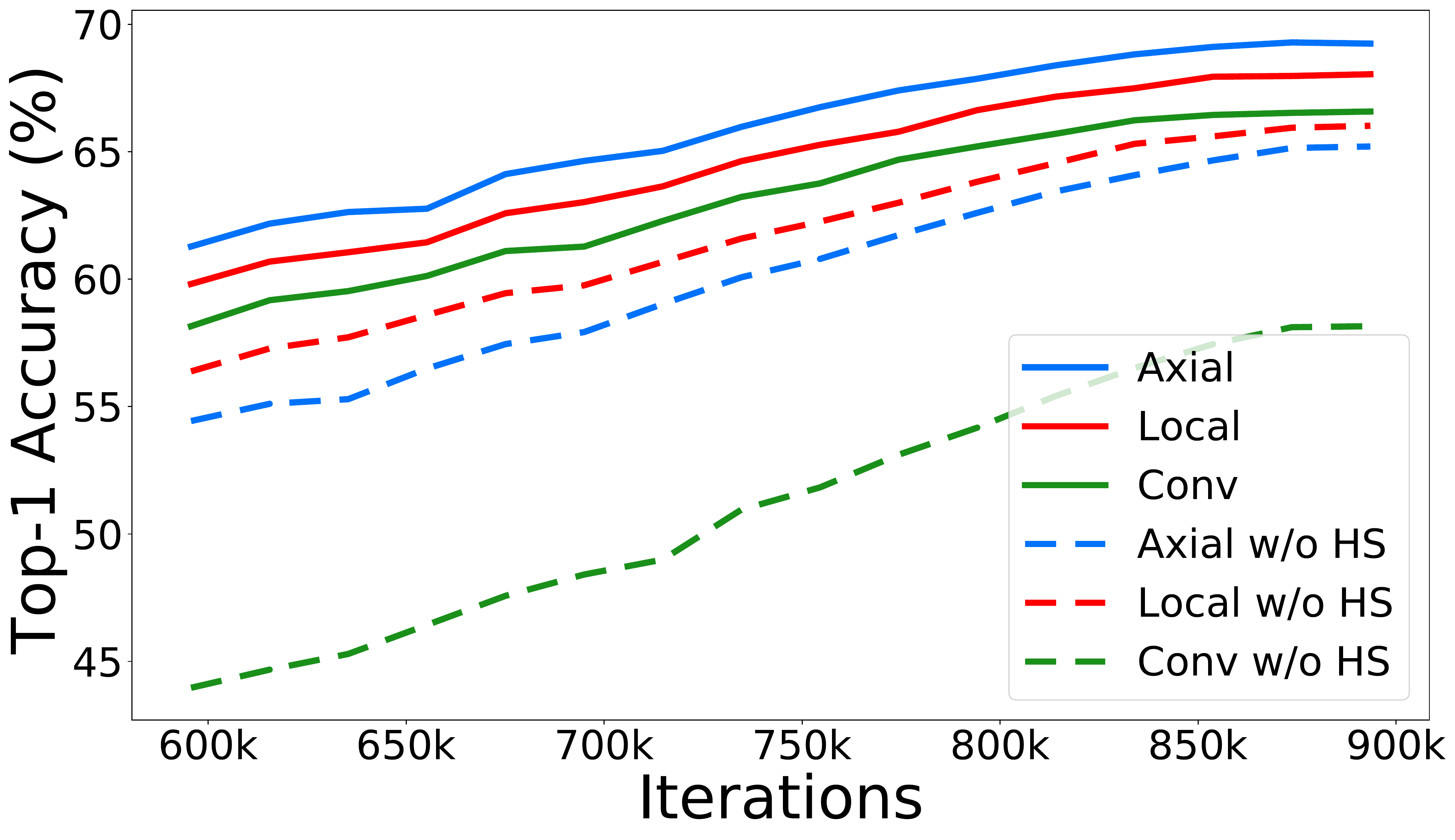}\quad \quad \quad &
\includegraphics[width=0.35\textwidth]{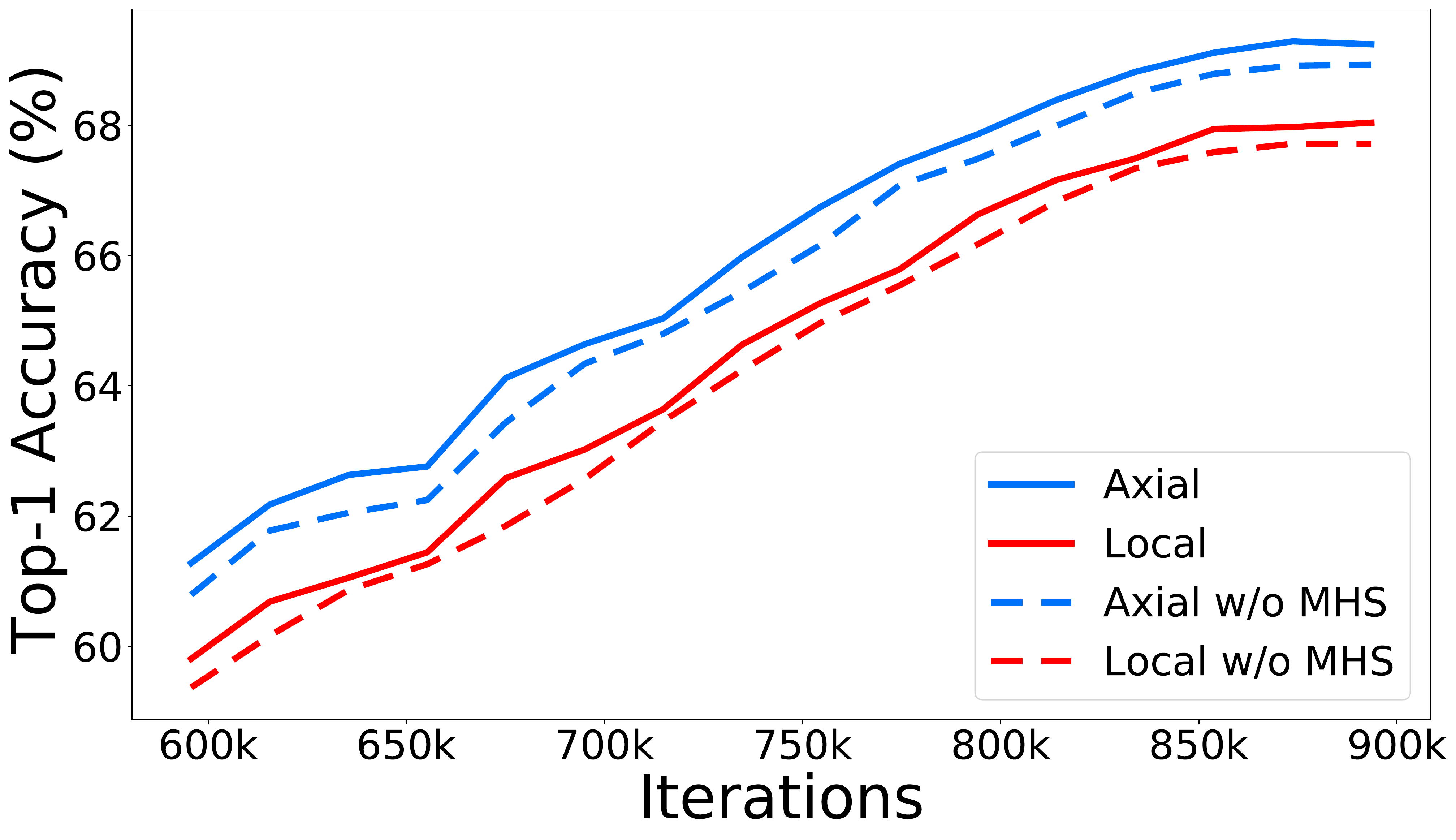}\vspace{-1ex}
\\(a) Effect of Hierarchical Sampling. & (b) Effect of Multi-Head Sharing. \\
\end{tabular}
\caption{Hierarchical Sampling and Multi-Head Sharing helps the training of the supernet.}
\vspace{-2ex}
\label{fig:dwssampling}
\end{figure}

\paragraph{Hierarchical Sampling.} \figref{fig:dwssampling} a) visualizes the supernet training curves with and without our proposed Hierarchical Sampling. We observe that sampling all candidates uniformly hurts convolution models a lot due to the low probability of convolution being sampled. However, with our proposed Hierarchical Sampling, all the model results are improved. It is worth mentioning that with our Hierarchical Sampling strategy, the local self-attention~\cite{parmar2019stand} models are improved too, even if they are not sampled as often as the case without Hierarchical Sampling. We hypothesize that the comparable performances of all three parallel candidates help the optimization of all operators.

\paragraph{Multi-Head Sharing.} \figref{fig:dwssampling} b) compares the training process of the supernet with and without our proposed Multi-Head Sharing. We observe that our Multi-Head Sharing strategy helps large multi-head self-attention models achieve higher accuracy. Similar curves are observed for small models as well, though the curves are omitted in the figure. In addition to the training curves of the supernet, we also analyze the searched model results. As shown in \tabref{tab:abla}, adopting Multi-Head Sharing in the supernet training improves the searched architecture by 1.5\%, from 75.4\% to 76.9\%.

\begin{figure}[t]
\begin{tabular}{cc}
\quad \; \includegraphics[width=0.35\textwidth]{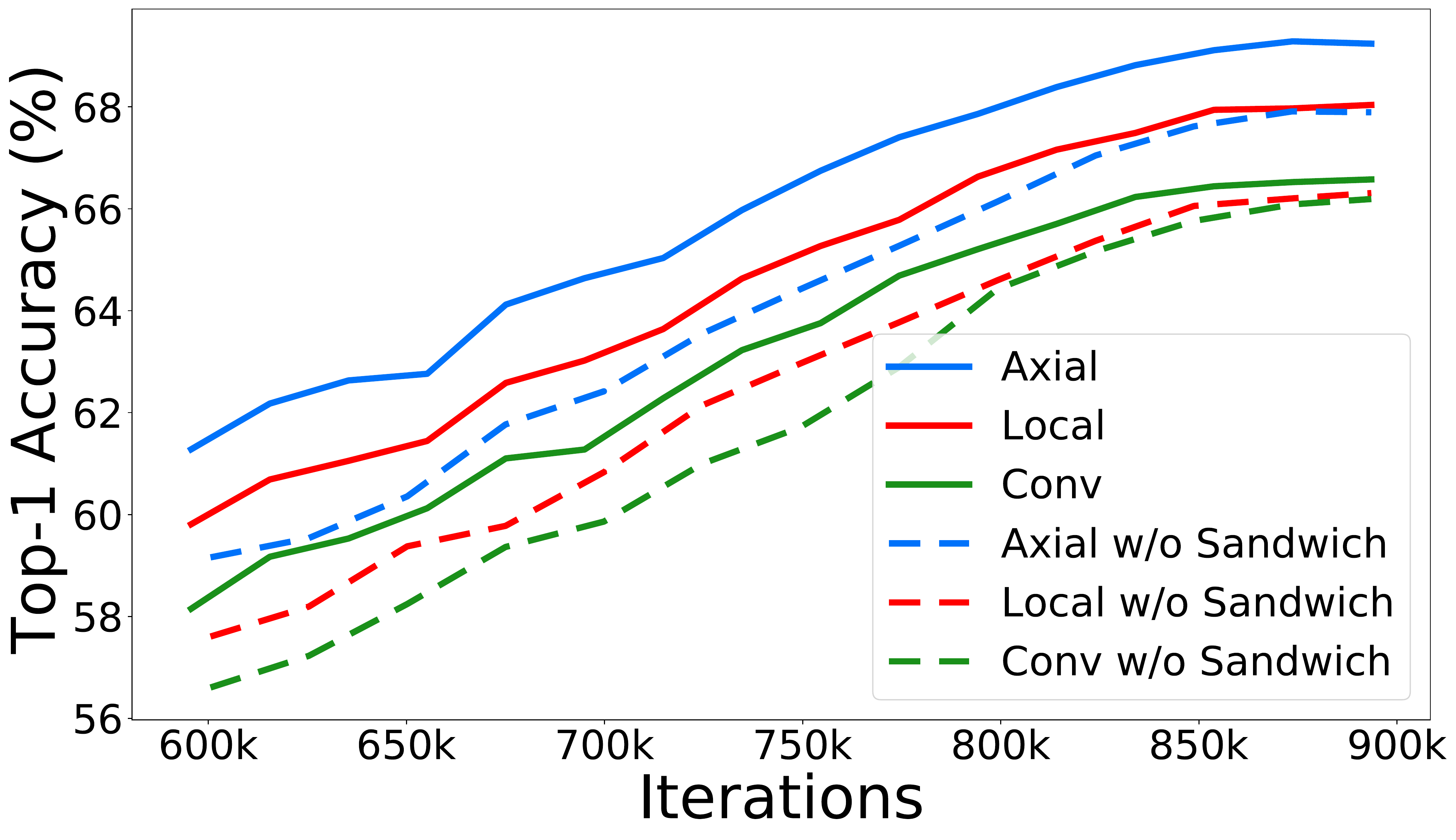} \quad \quad \quad &
\includegraphics[width=0.35\textwidth]{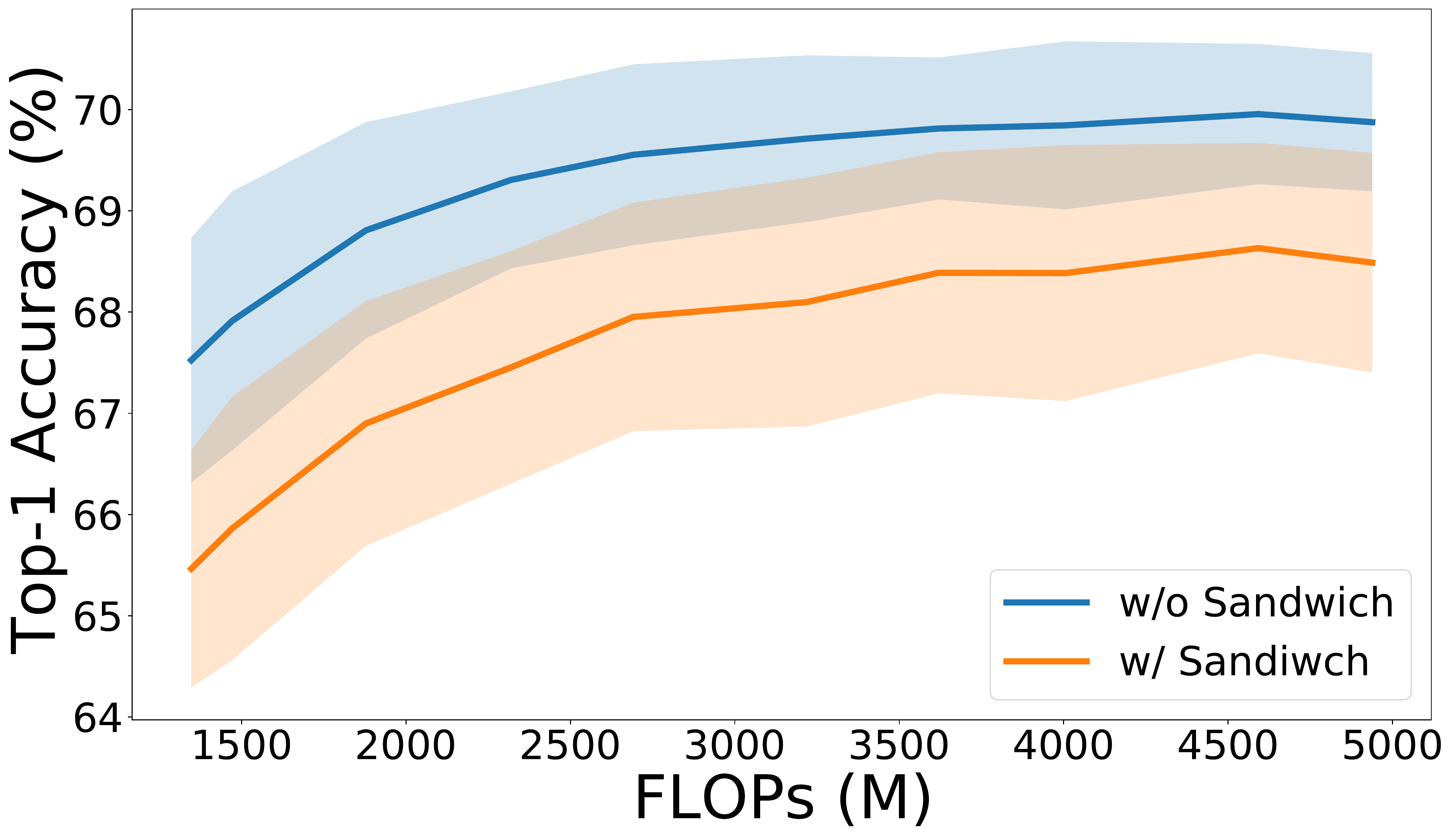}
\\(a) Better large models. & (b) Worse middle-sized models. \\
\end{tabular}
\caption{Effect of Sandwich rule to the supernet training.}
\label{fig:sand}
\vspace{-2ex}
\end{figure}

\paragraph{Sandwich rule.} \figref{fig:sand} a) plots the training curves with and without Sandwich rule~\cite{Yu2019UniversallySN, Yu2020BigNASSU}. It shows that Sandwich rule helps the training of large models by a large margin for all operators. \figref{fig:sand} b) shows the distribution of the sub-models selected from the supernet under different strategies. It can be observed that some middle-sized models are comparable with the biggest models because they are trained more and gain a lot from the large models optimization \cite{Yu2020BigNASSU}. However, this accuracy partial order cannot reflect the training from scratch accuracy, and the searched model accuracy in \tabref{tab:abla} also shows this. With Sandwich rule, the sampling distribution is changed and the ranking of these models are kept.

\begin{figure}[t]
\begin{tabular}{cc}
\quad \; \includegraphics[width=0.35\textwidth]{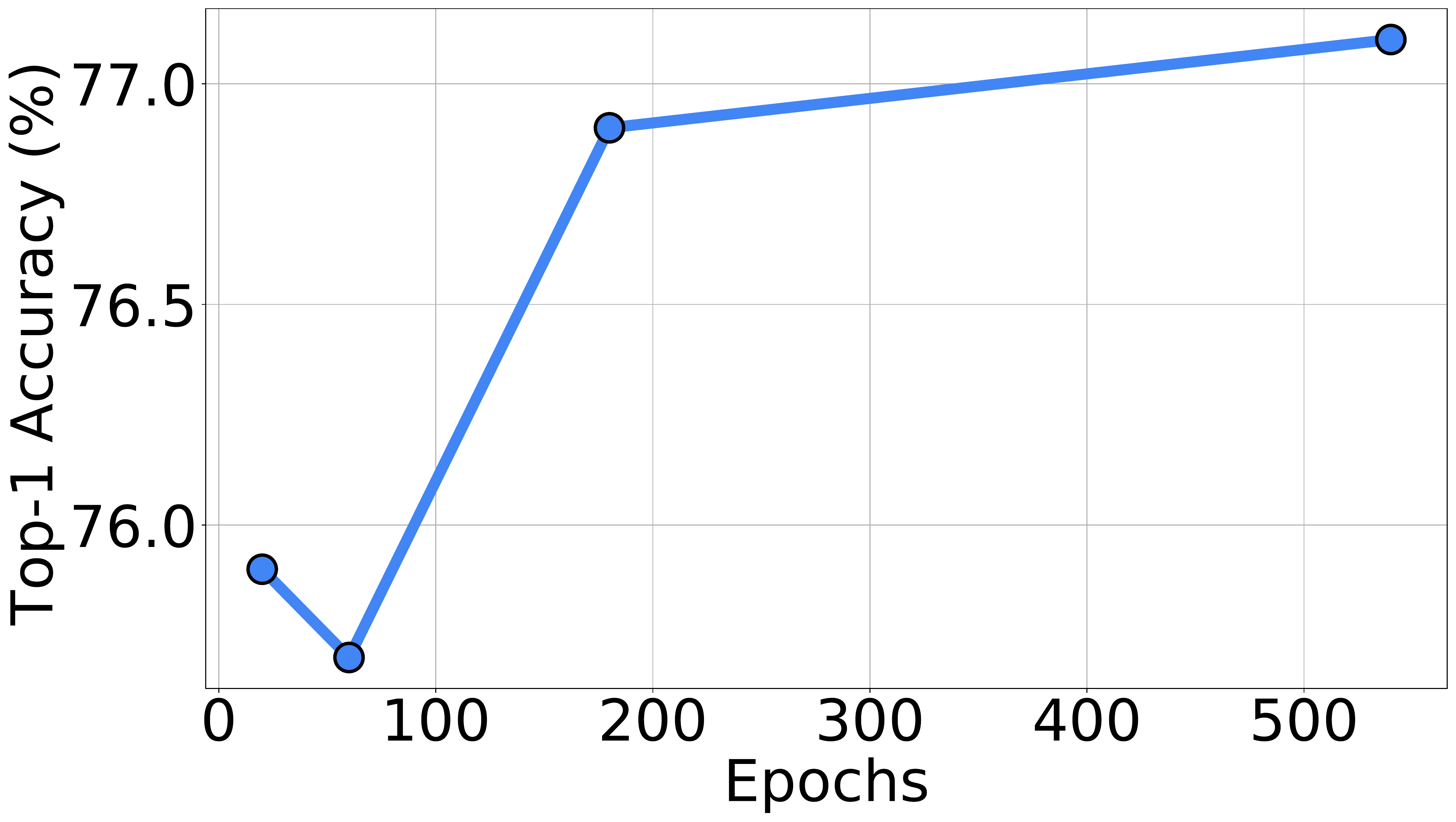} \quad\quad\quad &
\includegraphics[width=0.35\textwidth]{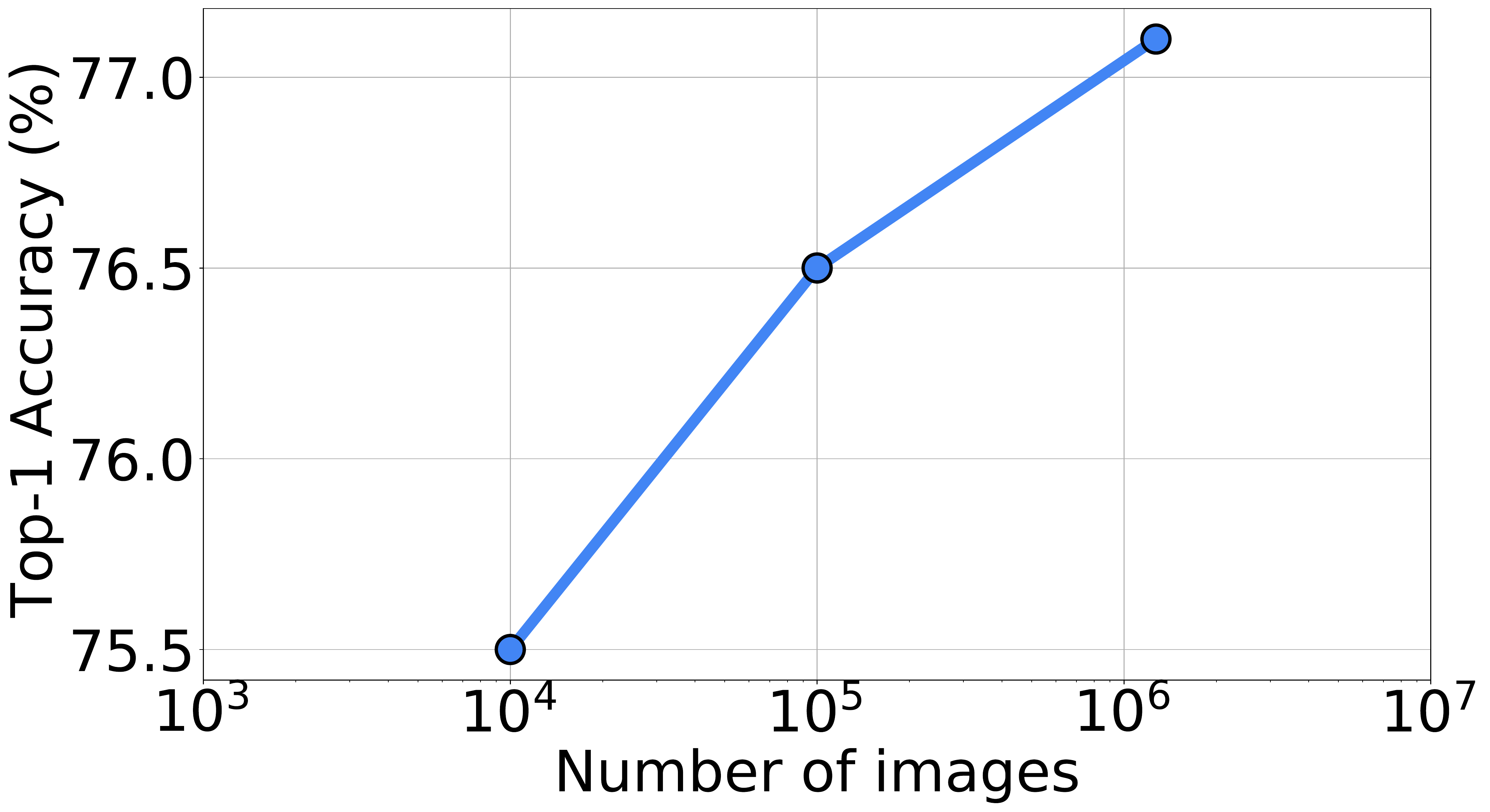}
\\(a) Varying number of epochs & (b) Varying number of images
\end{tabular}
\caption{TrioNet architectures become better if searched with more epochs and images.}
\label{fig:abla}
\vspace{-1.5ex}
\end{figure}

\paragraph{Number of epochs.} Weight-sharing NAS methods require a long supernet training schedule because the candidates need to be well-trained before they can accurately reflect how good each candidate is. Similarly, we evaluate how our TrioNet searching algorithm scales with longer training schedules. As shown in \figref{fig:abla} a), our searched models does not perform well with 20 epochs and 60 epochs, probably because the candidates have not been well-trained with such a short schedule. However, we do not observe a huge difference between 180 epochs and 540 epochs, probably because our supernet saturates with our simple ResNet-like training recipe and the weak data augmentation used.

\paragraph{Amount of data.}

Our TrioNet searching algorithm is also tested with different amount of data. To achieve this goal, we train the supernets with 10k, 100k images and the full training dataset (we still remove the evolutionary search set). Then, the searched  architectures are still trained from scratch on the full training set. In this way, we evaluate only the contribution of data on the searching algorithm, or the quality of the searched architecture, which is decoupled with the amount of data used to the searched models from scratch. \figref{fig:abla} b) shows that the searching on only 10k images gives a poor architecture, and our searching algorithm scales well, \ie, finds better architectures, with more data consumed.

\paragraph{Summary.}
\label{sec:ablaacc}
\begin{table*}[tb]
\footnotesize
    \centering
    \begin{tabular}{ccccc|cc}
        \hline
        \textbf{Settings} & \textbf{Sandwich} & \textbf{HS} & \textbf{MHS} & \textbf{Epochs} & \textbf{FLOPs (B)} & \textbf{Acc (\%)}\\
        \hline
        Random model & - & - & - & - & 4.8 & 74.7 \\
        \hline
        w/o Sandwich & & \checkmark & \checkmark & 180 & 3.6 & 74.9 \\
        w/o HS & \checkmark & & \checkmark & 180 & 3.5 & 75.6 \\
        w/o MHS & \checkmark & \checkmark & & 180 & 5.1 & 75.4 \\
        Ours & \checkmark & \checkmark & \checkmark & 180 & 5.2 & 76.9 \\
        \hline
        Ours w/ more epochs & \checkmark & \checkmark & \checkmark & 540 & 4.7 & \textbf{77.1} \\
        \hline
    \end{tabular}
    \caption{Comparison of searched models under different searching settings.}
    \label{tab:abla}
\vspace{-4ex}
\end{table*}

\tabref{tab:abla} summarizes the searched model accuracies of our ablated settings. We notice that without Sandwich rules \cite{Yu2019UniversallySN, Yu2020BigNASSU}, the searched model only achieves 74.9\% accuracy with 3.6B FLOPs, which is a middle-sized model. This indicates that the sandwich rules prevent our searching from biasing towards middle-sized models. Besides, our proposed Hierarchical Sampling shows 1.3\% performance gain on the searched model and Multi-Head Sharing strategy promotes the searched model with 1.5\% accuracy improvement. Furthermore, we also test a random model with the comparable size as the searched model, which only gets 74.7\% accuracy. Our NAS pipeline achieves 2.4\% improvement compared with the random model.




\vspace{-3ex}
\subsection{Results on Small Datasets}
\vspace{-1ex}

The goal of NAS is to automate the architecture design on the target data, task, and computation budget. From this perspective, we adapt our TrioNet NAS algorithm to other datasets beyond ImageNet~\cite{krizhevsky2012imagenet}, in order to evaluate if the algorithm is able to find a good architecture on the target datasets. These adaptation experiments are performed on five small datasets: Stanford Cars \cite{stanfordcars}, FGVC Aircraft \cite{fgvc}, CUB \cite{cub}, Caltech-101 \cite{caltech101} and 102 Flowers \cite{102flowers}. 

\paragraph{Overall Performance.}

\begin{table*}[tb]
\vspace{-2ex}
\footnotesize
    \centering
    \begin{tabular}{c|cc|cc|cc|cc}
        \hline
        \multirow{3}{*}{\textbf{Dataset}} & \multicolumn{2}{c|}{\textbf{ResNet-18}} & \multicolumn{2}{c|}{\textbf{Local-26}} & \multicolumn{2}{c|}{\textbf{Axial-26}} & \multicolumn{2}{c}{\textbf{TrioNet}} \\
         & \textbf{FLOPs} & \textbf{Acc} & \textbf{FLOPs} & \textbf{Acc} & \textbf{FLOPs} & \textbf{Acc} & \textbf{FLOPs} & \textbf{Acc} \\
         & (B) & (\%) & (B) & (\%) & (B) & (\%) & (B) & (\%) \\
        \hline
        Stanford Cars & 3.6 & 86.8 & 4.5 & 84.4 & \textbf{3.5} & 78.2 & 3.7 & \textbf{88.5} \\
        FGVC Aircraft & 3.6 & 79.8 & 4.5 & \textbf{82.4} & 3.5 & 74.4 & \textbf{2.9} & 82.1 \\
        CUB & 3.6 & \textbf{69.3} & 4.5 & 65.2 & 3.5 & 58.9 & \textbf{2.4} & 68.4 \\
        Caltech-101 & 3.6 & \textbf{75.3} & 4.5 & 71.4 & 3.5 & 66.1 & \textbf{2.4} & 72.2 \\    
        102 Flowers & 3.6 & \textbf{91.4} & 4.5 & 87.7 & \textbf{3.5} & 81.9 & 3.6 & 90.3 \\
        \hline
        Average & 3.6 & \textbf{80.5} & 4.5 & 78.2 & 3.5 & 71.9 & \textbf{3.0} & 80.3 \\
        \hline
    \end{tabular}
    \caption{Comparison with stand-alone models on different datasets.}
    \label{tab:dataset}\vspace{-3ex}
\end{table*}

\begin{table*}[tb]
\vspace{-1ex}
\footnotesize
    \centering
    \setlength{\tabcolsep}{1.3mm}
    \begin{tabular}{c|ccc|ccc|ccc|c}
        \hline
        \textbf{Settings} & \multicolumn{3}{c|}{\textbf{ResNet-18}} & \multicolumn{3}{c|}{\textbf{Local-26}} & \multicolumn{3}{c|}{\textbf{Axial-26}} & \textbf{TrioNet w/ supernet weights}\\
        \hline
        \textbf{Survival Prob} & 1.00 & 0.80 & 0.33 & 1.00 & 0.80 & 0.33 & 1.00 & 0.80 & 0.33 & -\\
        \hline
        \textbf{Stanford Cars} & 86.8 & 85.7 & 81.5 & 84.4 & 83.5 & 82.7 & 78.2 & 81.7 & 79.7 & \textbf{87.7} \\
        \textbf{FGVC Aircraft} & 79.8 & 78.6 & 73.8 & \textbf{82.4} & 75.8 & 72.9 & 74.4 & 74.0 & 74.6 & 80.5 \\
        \hline
    \end{tabular}
    \caption{Regularization effect of supernet training and stochastic depth.}
    \label{tab:joint}
\vspace{-2ex}
\end{table*}

\tabref{tab:dataset} shows the results of our TrioNet models searched directly on various small datasets \cite{stanfordcars, fgvc, cub, caltech101, 102flowers} and finetuned from supernet weights, as well as the accuracies of baseline models. Empirically, we notice that axial-attention~\cite{wang2020axial}, which outperforms ResNet~\cite{he2016deep} by a large margin on ImageNet~\cite{krizhevsky2012imagenet}, performs poorly on these small datasets with a big gap to ResNet (8.6\% accuracy on average), probably because the global axial-attention uses less induction bias than convolution or local-attention. However, in this challenging case, our TrioNet finds a better architecture than the hand-designed local self-attention~\cite{parmar2019stand} and axial self-attention~\cite{wang2020axial} methods. On these datasets, TrioNet is able to match the performance of the best operator, convolution in this case, with 20\% fewer FLOPs. This result, together with our ImageNet classification result, suggests that TrioNet robustly finds a good architecture no matter what operator the target data prefers.

\paragraph{Regularization Effect.}

Our sampling-based supernet training, where we sample an operator and then sample a candidate in the supernet, is similar to stochastic depth~\cite{huang2016deep} with a survival probability of 0.33 for each operator. Therefore, we compare our TrioNet (weights directly copied from the supernet) with stand-alone models of various survival probability on Stanford Cars \cite{stanfordcars} and FGVC Aircraft \cite{fgvc}, as shown in \tabref{tab:joint}. We notice that most stand-alone models perform worse with stochastic depth \cite{huang2016deep}. However, our TrioNet with its weights directly sampled from the supernet still outperforms these stand-alone models on Stanford Cars \cite{stanfordcars}. This suggests that the joint training of different operators is contributing to the performance as a better regularizer than stochastic depth \cite{huang2016deep}.

\vspace{-3ex}
\subsection{Results on Segmentation Tasks}
\vspace{-1ex}

\begin{table*}[tb]
\footnotesize
    \centering
    \begin{tabular}{c|cc|cccc}
        \hline
        \multirow{2}{*}{\textbf{Backbone}} & \multicolumn{2}{c|}{\textbf{Semantic Segmentation}} & \multicolumn{4}{c}{\textbf{Panoptic Segmentation}}\\
        & FLOPs & $mIoU$ & FLOPs & $PQ$ & $PQ^{Th}$ & $PQ^{St}$\\
        \hline
        ResNet-50 \cite{he2016deep} & 66.4B & \textbf{71.2} & 97.1B & 31.1 & 31.8 & 29.9 \\
        Axial-26 \cite{wang2020axial} & 33.9B & 67.5 & 60.4B & 30.6 & 30.9 & 30.2 \\
        TrioNet-B & \textbf{29.6B} & 69.3 & \textbf{52.3B} & \textbf{31.8} & \textbf{32.4} & \textbf{30.8} \\
        \hline
    \end{tabular}
    \caption{Comparison on segmentation tasks.}
    \label{tab:seg}
\vspace{-2ex}
\end{table*}

In this section, we evaluate TrioNet on semantic segmentation \cite{long2015fully, deeplabv12015} and panoptic segmentation \cite{kirillov2018panoptic} tasks. The ImageNet pretrained models are employed. For semantic segmentation, we apply DeepLabV3 \cite{chen2017deeplabv3} on PASCAL VOC datasets \cite{voc10}. For panoptic segmentation, we perform the experiments on COCO datasets \cite{lin2014microsoft} with Panoptic-DeepLab \cite{cheng2019panoptic} under a short training schedule. All experiments only replace the backbone with TrioNet-B.

\paragraph{Results.}

\tabref{tab:seg} shows the semantic segmentation and panoptic segmentation results. Using TrioNet-B as the backbone outperforms Axial-26 \cite{wang2020axial} by 1.8\% mIoU on semantic segmentation with 12.7\% less computation. On panoptic segmentation, TrioNet-B outperforms Axial-26 \cite{wang2020axial} by 1.2\% $PQ$ with 13.4\% fewer FLOPs.

\vspace{-3.5ex}
\section{Conclusion}
\vspace{-2ex}

In this paper, we design an algorithm to automatically discover optimal deep neural network architectures in a space that includes fully self-attention models~\cite{parmar2019stand, wang2020axial}. We found it is not trivial to extend the conventional NAS strategy \cite{Cai2020Once-for-All:, Yu2020BigNASSU} directly because of the difference between convolution and self-attention operators. We therefore specifically redesign the searching algorithm to make it effective to search for self-attention vision models. Despite our observation is from studying the self-attention module, we believe this is readily to extend to searching for other components such as normalization modules~\cite{ioffe2015batch}. 

\FloatBarrier

\bibliography{egbib}

\clearpage
\appendix

\section{Implementation Details}

\paragraph{Search Space}
Our search space contains four stages with $2, 3, 6, 3$ blocks in total. The selected number of blocks is $1, 2$ for the first stage, $2, 3$ for the second, $3, 4, 5, 6$ for the third and $1, 2, 3$ for the last stage. The other choices are shown in Tab. 1 in the main paper.

\paragraph{ImageNet}
Following the typical weight-sharing strategy, the original ImageNet training set is split into two subsets: 10k images for evolutionary search validation and the rest for training the supernet \cite{Cai2020Once-for-All:}. We report our results on the original validation set.

In our first stage of training, we apply SGD optimizer with learning rate $0.1$, Nesterov momentum $0.9$ and the weight decay $8e^{-5}$, which is only added to the largest model. Label smoothing~\cite{szegedy2016rethinking} is also adopted. We do not use dropout \cite{dropout} or color jitter since this training procedure is already strongly regularized. We train our supernet for $540$ epochs, where contains $10$ warmup epochs, with batchsize $32$ per gpu and $256$ in total. We use $\gamma=0$~\cite{goyal2017accurate} in the final BN~\cite{ioffe2015batch} layer for each residual block to stabilize the training procedure.

After searching, we train the searched model from scratch. We change the training epochs to $130$, and keep other hyper-parameters not changed.

For OFA \cite{Cai2020Once-for-All:} retraining, we apply the same training procedure on the models provided by their codebase. We employ the provided models with MACCs 0.6B, 0.9B, 1.2B and 1.8B (FLOPs are 1.2B, 1.8B, 2.4B and 3.6B). By using our setting that image input size is $224 \times 224$ and convert the stem as conv stem \cite{he2016deep, parmar2019stand}, these models FLOPs become 2.4B, 3.4B, 4.6B and 4.8B.

\paragraph{Small Datasets.}

We modify our ImageNet recipe by training 60k iterations with weight decay $1e^{-3}$, dropout 0.1, attention-dropout 0.1, and strong color jittering since the datasets are small and easy to overfit. For each dataset, 500 of the training images are selected for evolutionary search. After training the supernet and the evolutionary search, we directly sample the weights~\cite{Cai2020Once-for-All:,Yu2020BigNASSU} from the supernet and finetune the searched model for 1k iterations with a base learning rate $1e^{-3}$ with batchsize 128. We do not employ ImageNet~\cite{krizhevsky2012imagenet} pretraining in order to test the adaptation of our searching algorithm directly on the target data, instead of testing the transfer performance of the found architecture.

\paragraph{Segmentation}

For semantic segmentation on PASCAL VOC datasets \cite{voc10}, we use SGD optimizer with learning rate $0.01$, momentum $0.9$ and weight decay $5e^{-4}$. Models are trained with batchsize $16$ for $20K$ iterations and the input image size is $512\times512$. We use output stride $32$ and do not apply dilated operators in the backbone. For panoptic segmentation on COCO datasets, we apply Adam optimizer \cite{kingma2014adam} with learning rate $6.25e^{-5}$ and without weight decay. Models are trained with batchsize $8$ for $200K$ iterations and the input images size is $640\times640$.

\end{document}